\documentclass[conference]{IEEEtran}
\IEEEoverridecommandlockouts

\usepackage{cite}
\usepackage{amsmath,amssymb,amsfonts}
\usepackage{algorithmic}
\usepackage{graphicx}
\usepackage{makecell}
\usepackage{textcomp}
\usepackage{booktabs}
\usepackage{xcolor}
\usepackage{subfigure}
\def\BibTeX{{\rm B\kern-.05em{\sc i\kern-.025em b}\kern-.08em     T\kern-.1667em\lower.7ex\hbox{E}\kern-.125emX}}
\usepackage{balance}
\usepackage{hyperref}  
\hypersetup{
colorlinks=true,
linkcolor=blue,
urlcolor=blue,
citecolor=blue,
}

\begin{document}
\title{LimSim++: A Closed-Loop Platform for Deploying Multimodal LLMs in Autonomous Driving
\thanks{This work was supported by the Science and Technology Commission of Shanghai Municipality (Nos. 22DZ1100102 and 22YF1461400) and the National Key R\&D Program of China (No. 2022ZD0160104).}

\author{Daocheng Fu$^{1\ast}$, Wenjie Lei$^{21\ast}$, Licheng Wen$^{1}$, Pinlong Cai$^{1\dagger}$, Song Mao$^{1}$, Min Dou$^{1}$, Botian Shi$^{1\dagger}$, Yu Qiao$^{1}$

\thanks{$1$ Shanghai Artificial Intelligence Laboratory.}
\thanks{$2$ College of Control Science and Engineering, Zhejiang University.}
\thanks{${\ast}$ Equal contribution.} 
\thanks{$\dagger$ Corresponding author, Email: \{caipinlong, shibotian\}@pjlab.org.cn.}
}
}


\maketitle

\begin{abstract}
The emergence of Multimodal Large Language Models ((M)LLMs) has ushered in new avenues in artificial intelligence, particularly for autonomous driving by offering enhanced understanding and reasoning capabilities. This paper introduces LimSim++, an extended version of LimSim designed for the application of (M)LLMs in autonomous driving. Acknowledging the limitations of existing simulation platforms, LimSim++ addresses the need for a long-term closed-loop infrastructure supporting continuous learning and improved generalization in autonomous driving. The platform offers extended-duration, multi-scenario simulations, providing crucial information for (M)LLM-driven vehicles. Users can engage in prompt engineering, model evaluation, and framework enhancement, making LimSim++ a versatile tool for research and practice. This paper additionally introduces a baseline (M)LLM-driven framework, systematically validated through quantitative experiments across diverse scenarios. The open-source resources of LimSim++ are available at: \href{https://pjlab-adg.github.io/limsim-plus/}{https://pjlab-adg.github.io/limsim-plus/}.
\end{abstract}

\begin{IEEEkeywords}
Autonomous vehicle, closed-loop simulation, large language model, knowledge-driven agent

\end{IEEEkeywords}

\section{Introduction}

Recently, Multimodal Large Language Models ((M)LLMs) have sparked a significant wave in both theoretical research and practical applications within the artificial intelligence domain \cite{openai2023gpt4, touvron2023llama, chowdhery2022palm, gpt4v}. The capacity for generalized understanding and logical reasoning that (M)LLMs offer is particularly crucial in advancing trustworthy autonomous driving systems \cite{li2023knowledgedriven, cui2024survey, huang2023applications}. Numerous studies \cite{wen2023dilu, sharan2023llm, wang2023drivemlm, jin2023surrealdriver, ma2023lampilot} have explored this area using closed-loop simulation platforms like \textit{HighwayEnv} \cite{highway-env}, \textit{CARLA} \cite{dosovitskiy2017carla}, and \textit{NuPlan} \cite{caesar2021nuplan}, etc. Although these platforms facilitate the validation of (M)LLM's application potential, they are not without limitations, including unrealistic traffic flows, challenges in customizing scenarios, and a lack of adaptation specifically tailored for (M)LLM applications.

\begin{figure}
    \centering
     \includegraphics[width=1\linewidth]{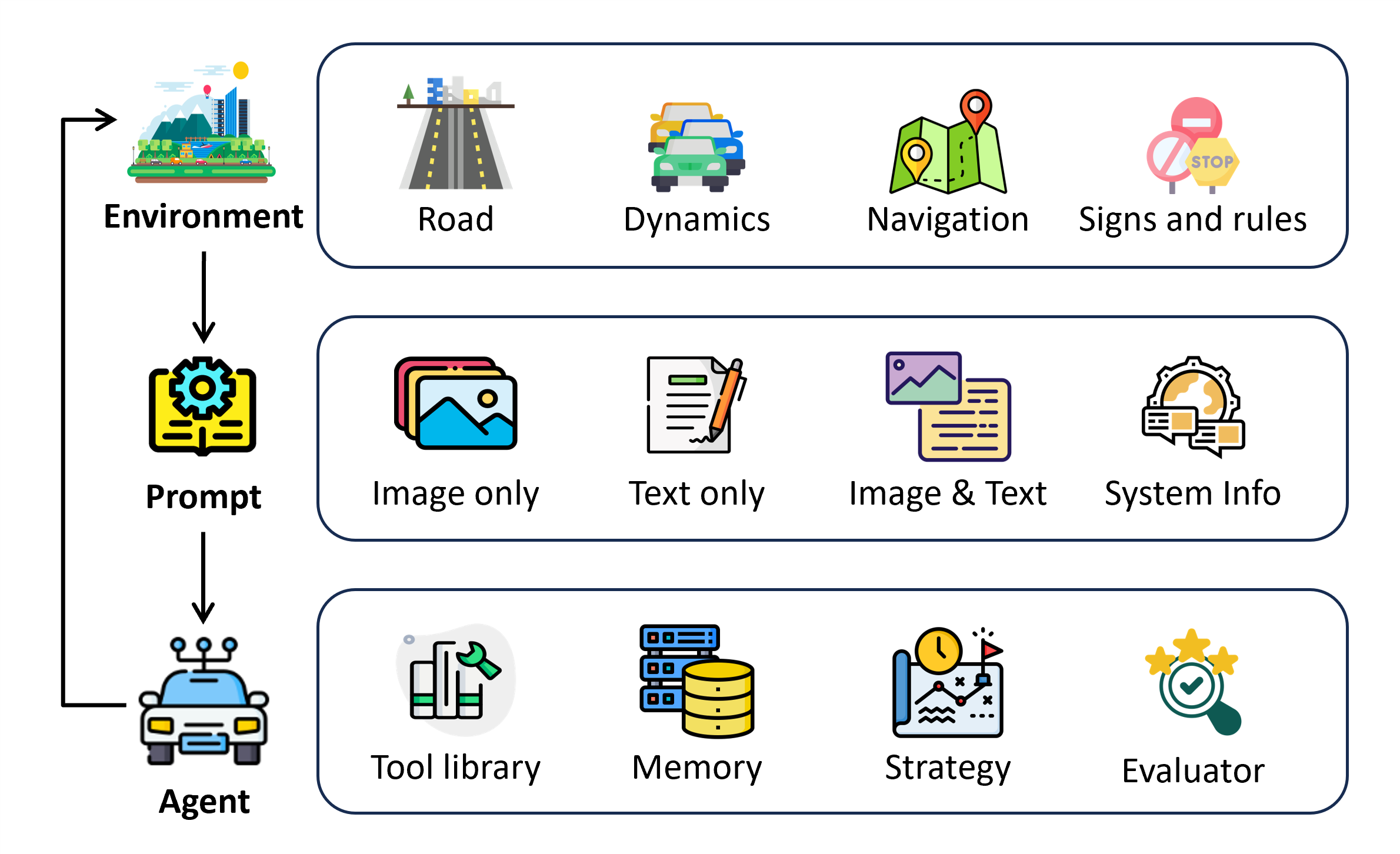}
      \caption{Platform composition. LimSim++ is the first closed-loop evaluation platform specifically developed for (M)LLM-driven autonomous driving.}
      \label{fig:introduction}
      \vspace{-15pt}
\end{figure}

Long-term closed-loop simulation platforms are considered an essential infrastructure for the ongoing iteration and advancement of current autonomous driving technologies \cite{li2023knowledgedriven, sharan2023llm, fu2024drive, shao2023lmdrive}, because these platforms are crucial for enabling autonomous driving models or systems to acquire and refine continuous learning capabilities, thereby improving their generalization abilities. 
Recognizing the limitations of existing simulation platforms, \textit{LimSim}  \cite{wen2023limsim} has been developed to facilitate extended-duration simulations across multiple scenarios with interactive capabilities. It incorporates advanced features such as collaborative decision-making among multiple vehicles, interactive replay, and comprehensive multidimensional scenario assessments. In this study, we attempt to develop an extended version of \textit{LimSim} tailored for (M)LLM applications, denoted as \textbf{LimSim++}. As illustrated in Fig.~\ref{fig:introduction}, LimSim++ provides a closed-loop system with road topology, dynamic traffic flow, navigation, traffic control, and other essential information. Prompts serve as the basis for the agent system supported by (M)LLM, incorporating real-time scenario information presented through images or textual descriptions. The (M)LLM-supported agent system exhibits capabilities such as information processing, tool usage, strategy formulation, and self-assessment.

Based on the developed platform, users can engage in the following research and practices: (1) \textbf{Prompt Engineering}: Users can create appropriate scenario descriptions and prompt cues for custom scenarios, facilitating the utilization of (M)LLMs for vehicle control. (2) \textbf{Model Evaluation:} Users can evaluate the performance of various (M)LLM-based models in autonomous driving within scenarios involving intricate interactive processes. (3) \textbf{Framework Enhancement:} Users can enhance submodules within the closed-loop framework we provide to achieve improved performance.

Our contributions can be summarized as follows:
\begin{itemize}
    \item \textbf{Introducing an open-source evaluation platform for (M)LLM in autonomous driving.} LimSim++ is the first provision of an open-source evaluation platform specifically designed for the research of autonomous driving with (M)LLMs, supporting scenario understanding, decision-making, and evaluation systems.
    \item \textbf{Proposing a baseline (M)LLM-driven closed-loop framework.} The baseline framework for realizing (M)LLM-driven autonomous driving comprises modules like multimodal prompt generation, decision-making, dynamic evaluation, reflection, memory, etc.
    \item \textbf{Validating the proposed platform through quantitative experiments across diverse scenarios.} Quantitative experiments designed for diverse scenarios validate the application potential of (M)LLM-driven vehicles in various scenarios, including intersections, roundabouts, and ramps. 
\end{itemize}

\section{Literature Review}

\subsection{(M)LLM-driven framework}

The advancements in (M)LLMs have garnered significant attention for exploring new paradigms in autonomous driving across scenario understanding, decision-making, and end-to-end mechanisms. Decision-making methods driven by Large Language Models (LLMs) demonstrate capabilities in overcoming the interpretability and generalization limitations of learning-based systems. \textit{Drive as You Speak} \cite{cui2024drive} integrated LLM's language and reasoning abilities into autonomous vehicles, achieving interactive design based on human-machine dialogue, which was also validated through real-world experiments \cite{cui2023large}. \textit{LanguageMPC} \cite{sha2023languagempc} utilized the LLM to identify crucial interacting vehicles for action guidance, generating observation matrix, weight matrix, and action bias to align with MPC-based vehicle control. \textit{DiLu} \cite{wen2023dilu} designed a closed-loop decision system based on the LLM, introducing memory and reflection models for performance improvement. \textit{GPT-Driver} \cite{mao2023gpt} tokenized input texts and combining them with an end-to-end architecture to achieve open-loop decision-making. \textit{SurrealDriver} \cite{jin2023surrealdriver} leveraged driving experiences from multiple drivers as chain-of-thought prompts, developing a coach agent module for human-like perception and feedback.

Moreover, (M)LLMs can handle image and video data for scenario understanding, aligning it with interpretable text to achieve integrated perception in decision-making. \textit{On the Road with GPT-4V(ision)} \cite{wen2023road} demonstrated the potential of Vision-Language Models (VLMs) in handling out-of-distribution scenarios, recognizing intent, and making wise decisions in actual driving environments. \textit{DriveMLM} \cite{wang2023drivemlm} employed a (M)LLM for the behavior planning module in an autonomous driving system, utilizing driving rules, user commands, and sensor inputs for decision-making and explanations. \textit{DME-Driver} \cite{han2024dme} combined pre-annotated \textit{GPT-4V (vision)} \cite{gpt4v} with manual corrections to simulate human driver logic, aligning logical decisions with precise control signals for autonomous vehicles, and trained specialized models based on the \textit{LLaVA} \cite{liu2023visual}. \textit{Dolphins} \cite{ma2023dolphins} integrated image content description, target position recognition, and logic reasoning for decision-making into a conversational driving assistant by employing the innovative grounded chain of thought.

\subsection{Dataset, Benchmark, and Platform}

Historical benchmarks for perception, prediction, and planning fall short of adequately validating the performance of (M)LLMs in autonomous driving, thereby failing to showcase the human-like understanding and reasoning abilities inherent in (M)LLMs. A potential solution involves a hybrid approach, combining manual annotation with automatic annotation with pre-trained models. Introducing innovative strategies that leverage natural language for comprehensive scenario understanding, logical reasoning, decision-making, and execution on diverse datasets, such as \textit{HDD} \cite{hdd} and \textit{nuScenes} \cite{nuscenes}, is both viable and promising.

The \textit{Honda Research Institute-Advice Dataset (HAD)}~\cite{had} comprised 5,675 driving video clips, featuring human-annotated textual advice describing the driver's actions and attention from the perspective of a driving instructor. 
An object-centric language prompt set and a new prompt-based driving task were introduced in \textit{NuPrompt} \cite{wu2023language} for driving scenes within 3D, multi-view, and multi-frame space, addressing this challenge by expanding the \textit{nuScenes} dataset with 35,367 language descriptions, averaging 5.3 object tracks each. \textit{Drive with LLMs} \cite{chen2023driving} presented a new dataset of 160k QA pairs derived from 10k driving scenarios, paired with high-quality control commands collected with reinforcement learning agent and question-answer pairs generated by teacher LLM, and then introduced an evaluation metric for driving QA and demonstrates our LLM-driver's proficiency in interpreting driving scenarios, answering questions, and decision-making. \textit{NuScenes-QA} \cite{qian2023nuscenes} comprised 34K visual scenes and 460K question-answer pairs, utilizing existing 3D detection annotations to generate scene graphs and employing manual design of question templates, followed by programmatically generating question-answer pairs. \textit{DriveLM} \cite{sima2023drivelm} introduced datasets using \textit{nuScenes} and \textit{CARLA}, presenting a VLM-based baseline approach that concurrently addresses Graph VQA and end-to-end driving. The experiments showcased Graph VQA as a simple and principled framework for scene reasoning.

While offline datasets contribute to refining the general capabilities of (M)LLMs for autonomous driving through fine-tuning, validating the model's adaptability across diverse scenarios remains a formidable challenge. The knowledge-driven paradigm emerges as a promising direction for realizing autonomous driving \cite{li2023knowledgedriven}, with its continuous learning hinging on continuous feedback within a closed-loop environment. Given the substantial cost associated with real-world testing \cite{yu2022autonomous}, closed-loop simulation testing becomes an essential facet of autonomous driving technology \cite{caesar2021nuplan, chen2023end, yang2023unisim}. This paper introduces LimSim++, designed to fulfill the research requirements of (M)LLM-driven autonomous driving. Tailored for end-to-end autonomous driving solutions, LimSim++ incorporates a unique combination of text and image prompt engineering and includes an optional module featuring a flexible and user-friendly driver agent.

\section{System Overview}

\begin{figure*}[!t]
    \centering
    \vspace{-5pt}    \includegraphics[width=0.75\linewidth]{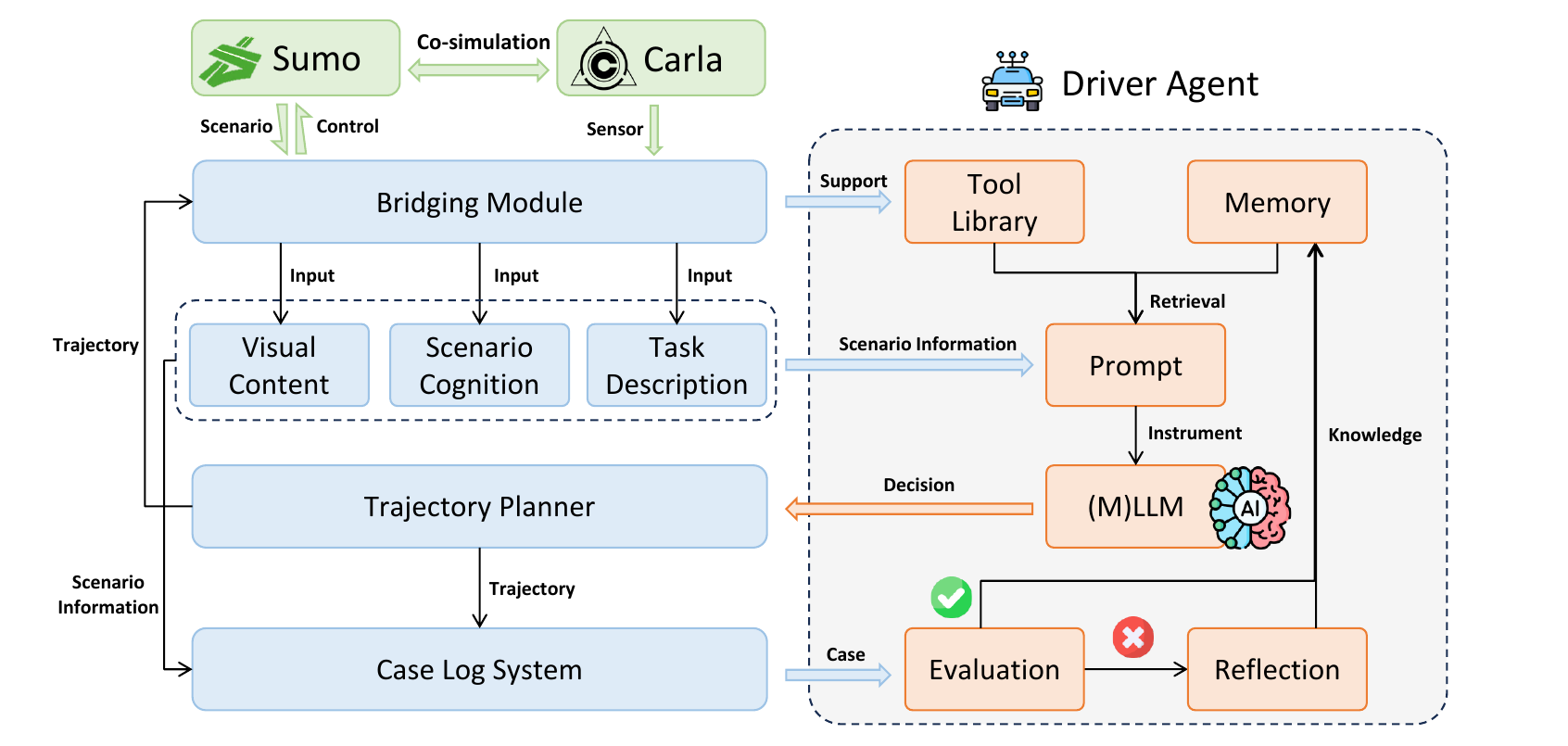}
      \vspace{-5pt}
      \caption{Framework of LimSim++. (1) Information Integration: Scenarios provided by \textit{SUMO} and visual contents from \textit{CARLA} are integrated into LimSim++ through the bridging module. (2) Prompt Engine: Constructing multimodal prompts to support (M)LLMs for understanding scenarios and tasks, including VLMs and LLMs. (3) Continuous Learning: The driver agent, driven by (M)LLM, makes behavioral decisions and continually enhances decision-making capabilities through mechanisms such as evaluation, reflection, memory, and tool library.}
      \label{fig:framework}
      \vspace{-10pt}
\end{figure*}

\subsection{Framework} 

LimSim++ facilitates the closed-loop simulation of (M)LLMs-driven autonomous driving, creating an environment for continuous learning. Illustrated in Fig.~\ref{fig:framework}, LimSim++ comprises two main components. The left side features the simulation system, extracting scenario information from \textit{SUMO} and \textit{CARLA}, packaging it into proper inputs for (M)LLMs (including visual content, scenario cognition, and task description). Simultaneously, the simulation system can plan trajectories based on (M)LLMs' decisions, achieving closed-loop simulation. On the right side is the (M)LLMs-powered driver agent, capable of interacting with the simulation system and making driving decisions. Within the driver agent, (M)LLMs can either directly interpret the driving scenario using the simulation system's input or perceive the environment by invoking tools. This information is formatted as natural language prompts, feeding into (M)LLMs for appropriate driving decisions. The reasoning and decision outputs from (M)LLMs influence vehicle actions and are stored in the case log system for knowledge accumulation. After simulation, decisions in the case log system are evaluated. Well-performing reasoning and decisions are directly added to memory, while those with poor performance are added after reflection. The memory assists (M)LLMs' driving decisions, enhancing the driver agent's performance.

\subsection{Multimodal Prompt}

LimSim++ offers various types and modalities of prompt inputs to meet the needs of diverse (M)LLMs for completing driving tasks. At each decision frame, LimSim++ extracts the road network and vehicle information around the ego vehicle. Then, this information of scenario descriptions and task descriptions is packaged and passed to the driver agent in natural language. LimSim++ has equipped the ego vehicle with six cameras based on the camera settings of \textit{nuScenes} \cite{nuscenes}. These cameras are capable of capturing panoramic images of the ego vehicle. 
The scenario description of LimSim++ is modular, providing real-time status, navigation information, and task descriptions. Users can freely combine scenario information based on the needs of the driver agent and package it into suitable prompts. Besides the information provided by LimSim++, users can define their tool library, allowing (M)LLM to obtain more custom information by invoking tools and assisting the driver agent in making decisions.

\subsection{Reasoning \& Decision}

LimSim++ serves as a closed-loop simulation evaluation platform, supporting the inference and decision-making processes of the dependent (M)LLMs, which can be done through zero-shot or few-shot driving approaches. Zero-shot driving involves making judgments directly based on obtained prompts; however, the hallucinatory issues of (M)LLMs can lead to decision failures. For non-specialized (M)LLMs, few-shot learning proves pivotal, allowing these models to acquire sensible solutions for driving tasks through exposure to a limited set of instances, especially when diverse scenarios necessitate distinct reactions. Undoubtedly, the prospect of leveraging pre-trained foundation models for autonomous driving is highly promising.

LimSim++ can handle various control signals derived from decision-making outcomes. If the driver agent provides only behavioral primitives, such as acceleration, deceleration, left turn, right turn, etc., LimSim++ offers control interfaces for these primitives, facilitating the conversion of the driver agent's decisions into vehicle trajectories. Furthermore, LimSim++ supports direct utilization of trajectories output by the driver agent to control vehicle motion, albeit with increased performance requirements on (M)LLMs.

\subsection{Evaluation}
\label{subsection:evaluation}

The evaluation module quantifies and assesses the vehicle behavior decisions made by the driver agent based on the analysis of vehicle trajectories. This process is an essential component within the continuous learning framework. The period of evaluation corresponds to the time interval between two consecutive decisions, determined by the decision frequency of (M)LLMs. The driving performance comprehensively considers route completion $R$ and driving score $S$. The route completion $R$ indicates the ratio of completed route length to total length, shown as
\begin{equation}
R=L_{completed}/L_{total},
\label{eq: r}
\end{equation}
where $L_{completed}$ represents the length of the route completed by the driver agent, and $L_{total}$ represents the total length of the preset route.
The driving score $S$ encompasses various factors, shown as
\begin{equation}
S = \alpha^{\lambda_1} \beta^{\lambda_2} \gamma^{\lambda_3} (k_1  r_{c} + k_2  r_{e} + k_3  r_{s}),
\label{eq: s}
\end{equation}
where three penalty multiplier terms, $\alpha$, $\beta$, and $\gamma$, are denoted for collisions, signal violations, and speed violations. $\lambda_1$, $\lambda_2$, and $\lambda_3$ represent the occurrence times of three types of punished behaviors in the whole trajectory. $k_1$, $k_2$, and $k_3$ are weighted coefficients for trajectory quality evaluation for each decision, incorporating ride comfort (denoted as $r_c$), driving efficiency (denoted as $r_e$), and driving safety (denoted as $r_s$), defined as below.

1) \textbf{Ride comfort}. Lateral and longitudinal accelerations, as well as jerking, significantly influence ride comfort, emphasizing the crucial need to constrain these factors within limited ranges. The calculation of ride comfort is presented as
\begin{equation}
    r_{c} = \left(s_{x_{a}}+s_{x_{j}}+s_{y_{a}}+s_{y_{j}}\right)/4,
\end{equation}
where $s_{x_a}$, $s_{x_j}$, $s_{y_a}$, and $s_{y_j}$ represent the reference values for comfort regarding lateral acceleration, lateral jerk, longitudinal acceleration, and longitudinal jerk. Utilizing empirical data, reference values can be established for various driving styles (cautious, normal, and aggressive), and specific comfort scores can be derived using interpolation methods~\cite{bae2020self}.

2) \textbf{Driving efficiency}. Velocity is pivotal for driving efficiency. In regular traffic conditions, vehicles should strive to maintain a velocity at least equivalent to the average velocity of surrounding vehicles. In sparse traffic conditions, vehicles should aim to closely approach the road's velocity limit to ensure driving efficiency. The calculation of driving efficiency is presented as
\begin{equation}
r_{e}= \begin{cases}
1.0 ,& \textit{if } v_e \geq v^*\\ 
v_e/v*, & \textit{else} \\ 
\end{cases},
\end{equation}
where $v_e$ represents the velocity of the ego vehicle. $v^*\in\{v_{avg}, v_{limit}\}$, where $v_{avg}$ and $v_{limit}$ are the average velocity of surrounding vehicles and the velocity limit, respectively.

3) \textbf{Driving safety}. The risk level is commonly assessed through Time to Conflict (TTC). When TTC falls below a specified threshold, it signifies the presence of potential risks that warrant penalties. The calculation of driving safety is presented as
\begin{equation}
r_{s} = \begin{cases}
1.0 , & \textit{if } \tau_e \geq \tau_{threshold} \\ 
\tau_e/\tau_{threshold}, & \textit{else} \\ 
\end{cases},
\end{equation}
where $\tau_e$ is the TTC of the ego vehicle, and $\tau_{threshold}$ is the threshold of TTC.

Due to the different execution periods of various vehicle behaviors, such as lane-changing behaviors having much longer execution periods than acceleration or deceleration behaviors, evaluating decision strategies based on fixed-length trajectory segments requires modification.
Utilizing the driving score derived from the whole trajectories, it is possible to comprehensively consider consecutive trajectory segments around low-scoring driving scores to assess the performances in making decisions regarding complex vehicle behaviors. However, unavoidably, for planners with high-frequency decision-making, some erroneous decisions may be corrected by subsequent decisions, rendering the driving score ineffective in detecting failures. 
Evaluating the rationality of current vehicle decisions based on predictions of future scenarios, particularly the motion predictions of surrounding vehicle trajectories, appears to be promising, such as the idea of \textit{Driving into the Future}~\cite{wang2023driving}. 

\subsection{Reflection \& Memory}

LimSim++ introduces a reflection and memory mechanism for continuous learning, aiming to enhance the driver agent's performance. Upon completing a segment of the path, the evaluator assesses the driving decision. LimSim++ utilizes these evaluation results for continuous learning, as they closely correlate with the decision-making performance of the driver agent. Decisions with high scores are directly integrated into the memory module. For decisions with low scores, LimSim++ prompts (M)LLMs to engage in self-reflection or error correction to refine decision accuracy, often with the assistance of human experts. Subsequently, the refined reasoning outcomes are integrated into the memory module.

Specifically, during self-reflection, the driver agent identifies areas for improvement based on evaluator scores and adjusts these aspects to optimize the reasoning process. In cases where the driver agent lacks reflection capacity, expert error correction is employed to ensure correct reasoning. The refined reasoning outcomes will act as few-shot instances to aid (M)LLM in subsequent decision-making processes when the driver agent encounters analogous scenarios. LimSim++ employs a vector database for the storage and retrieval of similar scenarios \cite{wen2023dilu}.

\section{Experiment}

\subsection{Experiment Setup}

LimSim++ offers diverse driving scenarios, enables users to customize paths, and facilitates dynamic, large-scale, long-duration simulation for interactive learning. Moreover, LimSim++ presents a continuous learning framework to enhance the application and development of (M)LLMs in autonomous driving. In the experiment, we establish the success criterion as the vehicle reaching the specified destination. Failure occurs if there is a collision, driving in the wrong lane, or exceeding the time limit, leading to the direct termination of the simulation. A time limit of 80 seconds is set for each scenario. When assessing the driving process, the parameters in the evaluation module are configured as follows: $\tau_{threshold}=5.0$, $\alpha=0.6$, $\beta=0.7$, $\gamma=0.9^{10/\epsilon}$, where $\epsilon$ is the times of decision-making in the whole simulation. In the continuous learning framework, we employ \textit{Chroma DB}\footnote{\url{https://www.trychroma.com/}} as the vector database for memory storage. 
The scenario descriptions are transformed into vectors using \textit{OpenAI}'s embedding tools of \textit{text-embedding-ada-002}\footnote{\url{https://platform.openai.com/docs/guides/embeddings}}.

In this section, we outline various experimental setups and provide baseline results. Initially, we assess the performance of (M)LLMs in comprehending environmental information and executing human verbal commands across different driving scenarios. Subsequently, we evaluate the driving capabilities of (M)LLMs in a zero-shot setting without prior training. Lastly, we demonstrate the efficacy of memory-powered continuous learning methods in enhancing the driving proficiency of (M)LLMs.
The demonstrations of the experiments are available at \href{https://pjlab-adg.github.io/limsim-plus/}{https://pjlab-adg.github.io/limsim-plus/}.

\subsection{Various Scenarios Supporting}

\begin{figure}
    \centering
    \includegraphics[width=1\linewidth]{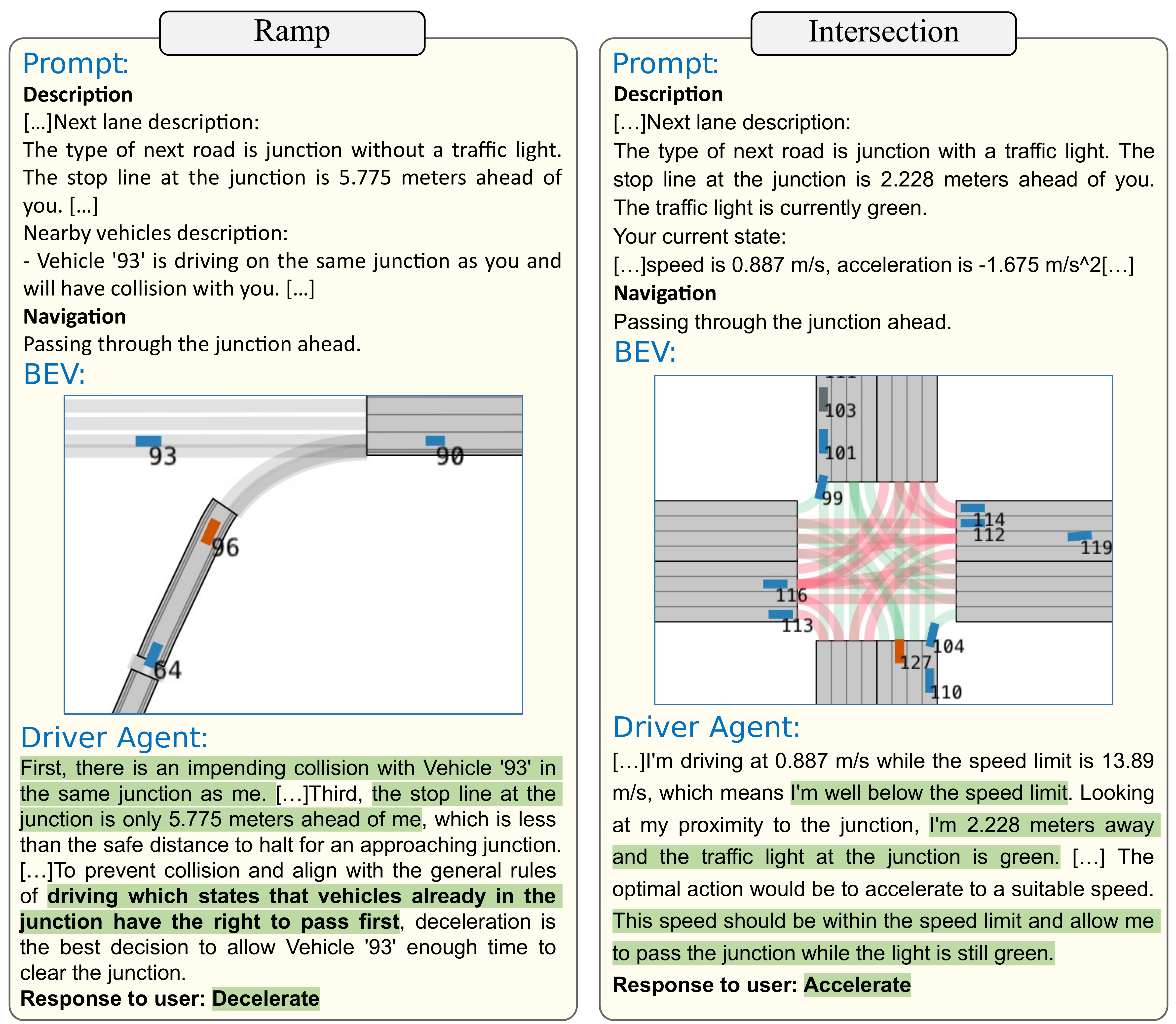}
    \caption{Prompts and driving decisions for critical scenarios. \colorbox[RGB]{189,218,165}{Green} highlights the right answer from the \textit{GPT-4}. }
    \label{fig:various_sce}
    \vspace{-15pt}
\end{figure}

LimSim++ can provide information output for various scenarios to support (M)LLMs in making driving decisions. 
We built a driving agent based on \textit{GPT-4}~\cite{openai2023gpt4} and let it undergo driving tests in the scenarios of intersections, roundabouts, and ramps, with each scenario being simulated 10 times. Table~\ref{tab:multi-scenarios} shows the performances in different scenarios, where route completion and driving score are detailed in Section~\ref{subsection:evaluation}. The average decision time refers to the average time it takes for the driver agent to make a decision each time. The longer the average decision time, the longer the response time of the (M)LLM, which will lead to a larger online simulation delay. The success rate represents the proportion of times passed in 10 simulation tests. As shown in Table~\ref{tab:multi-scenarios},  the driver agent consistently achieves over 90\% route completion and the driving scores are higher than 70 in all three scenarios. This suggests that (M)LLMs can make proper driving decisions according to crucial scenario information offered by LimSim++. 

\begin{table}[tb]
\centering
\caption{Multi-scenario Performance Validation}

\resizebox{\columnwidth}{!}{
\begin{tabular}{lccc}
\toprule
\textbf{Metric} & \textbf{Intersection} & \textbf{Roundabout}  & \textbf{Ramp}  \\
\midrule
\textbf{Route completion (\%)$\uparrow$}    
& 94.88 $\pm$ 16.18  & 93.34 $\pm$ 17.89 & 97.44 $\pm$ 8.08  \\
\textbf{Driving score} $\uparrow$ & 73.18 $\pm$ 10.54  & 71.05 $\pm$ 10.18   & 76.58 $\pm$ 13.42 \\
\textbf{Avg. decision time (s)} $\downarrow$ & 19.13 $\pm$ 5.41   & 25.60 $\pm$ 5.82    & 26.65 $\pm$ 7.34  \\
\textbf{Success rate} $\uparrow$  & 90\%  & 90\%  & 90\%              \\
\bottomrule
\end{tabular}}
\label{tab:multi-scenarios}
\vspace{-10pt}
\end {table}

Fig.~\ref{fig:various_sce} shows the reasoning process of the driver agent at a ramp and an intersection. The left figure demonstrates the scenario description and the decision process when the vehicle exits the ramp. The scenario description provides necessary information about the next lane and the surrounding vehicles. 
In this scenario, the ego vehicle on the ramp is going to merge into the main road, posing a collision risk with vehicle 93.
The driver agent analyzes the description and navigation information and finally concludes that it needs to yield to vehicle 93 and choose to decelerate. 
The right figure showcases the decision-making process that occurs when the traffic light transitions from red to green at the intersection. The driver agent chooses to accelerate based on the requirement to cross through the intersection quickly when the light is green. These two examples show that the driver agent powered by \textit{GPT-4} has strong common sense, showcasing its ability to perform effectively across diverse scenarios without training.

\begin{table}[t]
\centering
    \caption {Zero-shot driving with Different (M)LLMs}    
\resizebox{\columnwidth}{!}{
\begin{tabular}{lccc}
\toprule
\textbf{Metric} & \textbf{\textit{GPT-3.5}} & \textbf{\textit{GPT-4}}  & \textbf{\textit{GPT-4V}}   \\ 
\midrule
\textbf{Route completion (\%)}$\uparrow$          & {80.76 $\pm$ 24.44}  & \textbf{94.18 $\pm$ 18.42}   & 52.23 $\pm $25.62  \\
\textbf{Driving score} $\uparrow$                 & {64.74 $\pm$ 16.22}  & \textbf{78.24 $\pm$ 12.20}   & 51.93 $\pm $21.62 \\
\textbf{Avg. decision time (s)}$\downarrow$      & \textbf{2.43 $\pm$ 0.79}    & 15.97 $\pm$ 5.51    & 13.22 $\pm $3.82  \\
\textbf{Success rate}$\uparrow$                   & 50\%               & \textbf{90\%}              & 20\%              \\
\bottomrule
\end {tabular}}
\label{tab:zero-shot_driving}
      \vspace{-10pt}
\end {table}

\subsection{Comparison among different (M)LLMs}

LimSim++ supports different types of (M)LLMs for closed-loop driving simulation. We select a path in \textit{CARLA} \textit{Town06}\footnote{\url{https://carla.readthedocs.io/en/latest/map_town06/}} to perform closed-loop driving and compare the performances of different driver agents powered by \textit{GPT-3.5}, \textit{GPT-4}, and \textit{GPT-4V}, respectively. The agents with \textit{GPT-3.5} and \textit{GPT-4} utilize text descriptions as input, whereas another agent with \textit{GPT-4V} directly processes 2D camera images. 
Table~\ref{tab:zero-shot_driving} shows the performances of the three driver agents in closed-loop simulation. 
Among them, the driver agent powered by \textit{GPT-4} achieves the highest success rate of 90\% and the highest driving score of 78.24.
However, the decision time for \textit{GPT-4} tends to be longer on average, as it requires a substantial number of tokens to carry out an extensive reasoning process, resulting in increased delays. \textit{GPT-4V} can process camera images directly, whereas it struggles to infer dynamic information of surrounding vehicles from a single multi-view image time frame. This challenge significantly complicates its decision-making process, leading to a driving score of 51.93 and a route completion rate of 52.32\%, the lowest among the three agents evaluated.

\subsection{Enhancement with Memory Mechanism}

\begin{table}
\centering
\caption {Verification of Memory Mechanism}
\begin{tabular}{lcc}
\toprule
\textbf{Metric} & \makebox[0.13\textwidth][c]{\textbf{Without memory}} & \makebox[0.13\textwidth][c]{\textbf{With memory}} \\
\midrule
\textbf{Route completion (\%)} $\uparrow$     & 80.76 $\pm$ 24.44  & \textbf{100.00 $\pm$ 0.00} \\
\textbf{Driving score}     $\uparrow$         & 64.74 $\pm$ 16.22  & \textbf{89.22 $\pm$ 2.72}    \\
\textbf{Avg. decision time (s)}$\downarrow$  & \textbf{2.43 $\pm$ 0.79}    & 3.24 $\pm$ 0.65     \\
\textbf{Success rate}   $\uparrow$            & 50\%               & \textbf{100\%}              \\
\bottomrule
\end{tabular}
\label{tab:ablation_study}
\end {table}

\begin{figure}[t]
    \centering
    \includegraphics[width=0.8\linewidth]{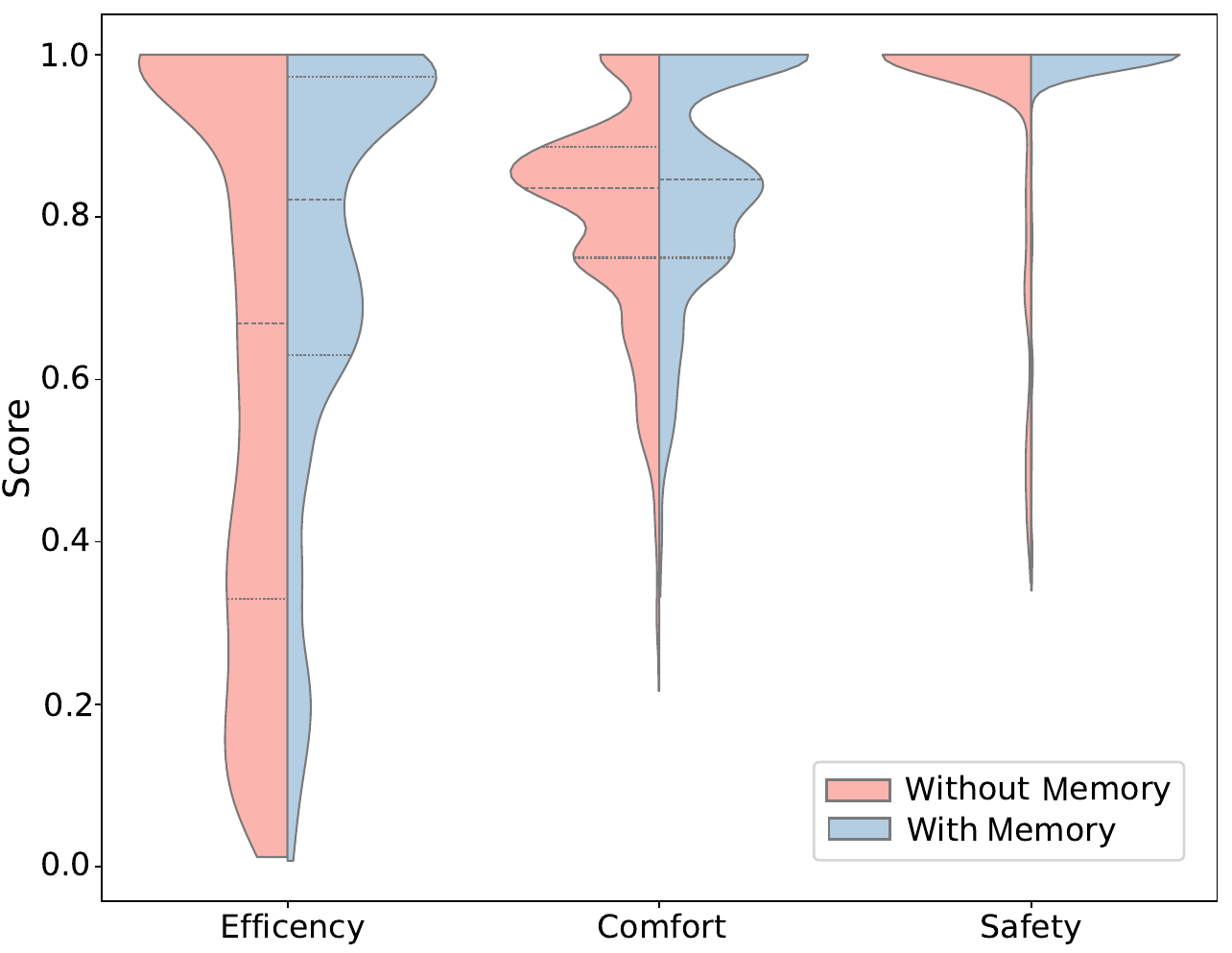}
    \caption{Comparison of driver agents with and without memory in terms of efficiency, comfort, and safety. Notably, the agent with memory uses 10 memory items with 3 shots.}
    \label{fig:ablation_vio}
    \vspace{-15pt}
\end{figure}

LimSim++ incorporates a continuous learning framework to improve the driving performance of (M)LLMs by expanding the reflection and memory modules. 
We create a driver agent powered by \textit{GPT-3.5} to verify the effectiveness of such a continuous learning framework. 
Building upon the results of the previous experiment, we refine the incorrect decision-making process with expert experience. The modified decision results are then added to the driver agent's memory modules, serving as driving experiences to provide few-shot instances for the agent in subsequent decision-making processes. In our experiments, three similar instances are retrieved by a scenario query method for few-shot driving.

Table~\ref{tab:ablation_study} and Fig.~\ref{fig:ablation_vio} show the driving performance of the driver agent with or without memory. Table~\ref{tab:ablation_study} shows that the success rate of the agent with memory can reach 100\%, and the driving score is also 37.8\% higher than the agent without memory. However, the agent with memory takes longer to make decisions because it needs to retrieve similar memory items. 
Fig.~\ref{fig:ablation_vio} shows the scores of the two driver agents on the three trajectory indicators of efficiency, comfort, and safety, which are introduced in detail in Section~\ref{subsection:evaluation}. The agent with memory performs significantly better than the agent without memory on three indicators. 

\begin{figure}
    \centering
    \includegraphics[width=1\linewidth]{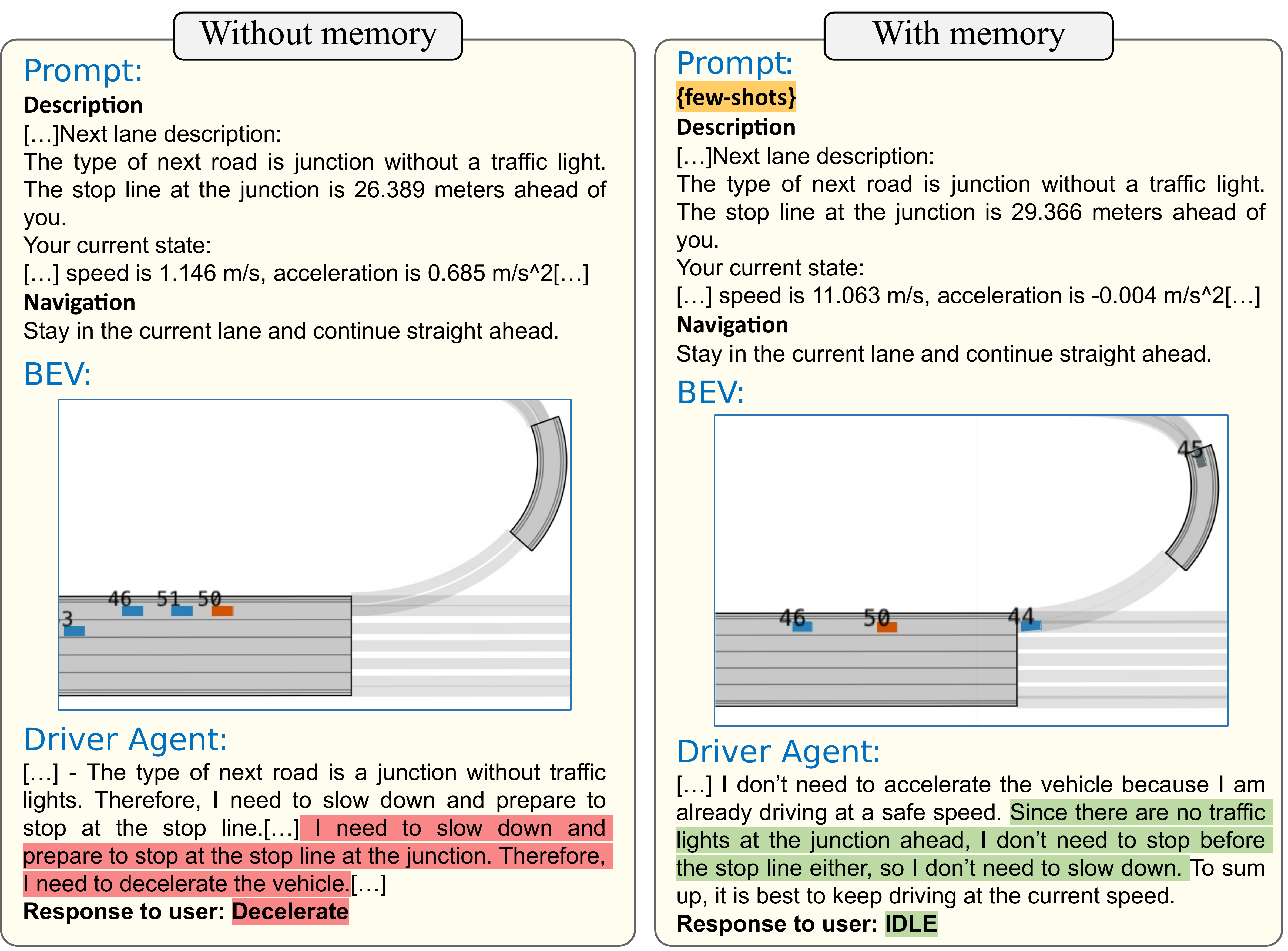}
    \caption{Enhancement with memory in an intersection. \colorbox[RGB]{189,218,165}{Green} highlights the right answer. \colorbox[RGB]{250,156,154}{Red} highlights the wrong answer. \colorbox[RGB]{255,204,102}{Yellow} highlights the similar experience drawn from the memory module, which includes the past scenario descriptions and correct reasoning processes.}
    \label{fig:ablation_1}
    \vspace{-10pt}
\end{figure}

\begin{figure}
    \centering
    \includegraphics[width= 1\linewidth]{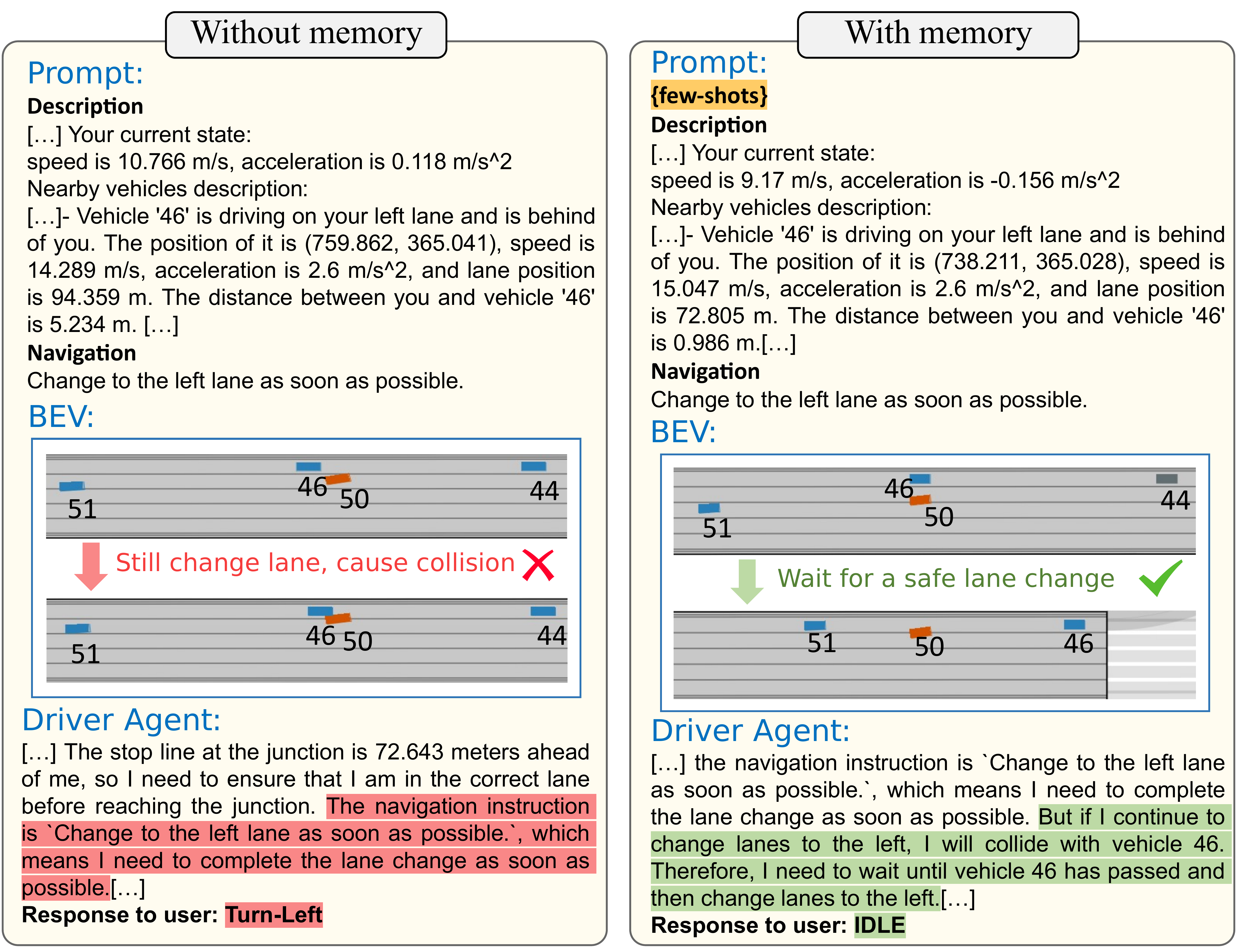}
    \caption{Enhancement with memory for a lane-changing process. \colorbox[RGB]{189,218,165}{Green} highlights the right answer. \colorbox[RGB]{250,156,154}{Red} highlights the wrong answer.\colorbox[RGB]{255,204,102}{Yellow} highlights the similar experience drawn from the memory module, which includes the past scene descriptions and correct reasoning processes.}
    \label{fig:ablation_2}
    \vspace{-15pt}
\end{figure}

To better demonstrate the impact of memory on the driver agent, two critical scenarios are selected for further analysis. 
Fig.~\ref{fig:ablation_1} shows the different decisions made by two agents when approaching the intersection. 
In the scenario where the agent lacks memory capabilities, the driver agent assumes it must halt at the stop line. Consequently, it stops for an extended period at a considerable distance from the intersection, ultimately causing the experiment to fail due to a timeout.
With the guidance of memory, the driver agent accurately assesses that it does not need to stop at the intersection without traffic lights, and thus reaches the destination smoothly. 

Fig.~\ref{fig:ablation_2} illustrates distinct decisions made by the agent in scenarios with elevated lane-changing risk. The driver agent lacking memory tends to prioritize lane changes, overlooking the potential collision risk with vehicle 46, ultimately resulting in an accident. Conversely, the driver agent with a memory module recognizes the substantial lane-changing risk and chooses to maintain a safe distance behind vehicle 46 before executing the lane-changing motion. Thus, the inclusion of memory is beneficial, rendering the driver agent's reasoning process more judicious.

\section{Conclusion}
The paper proposes an integrated platform called LimSim++ for scenario understanding, decision-making, and evaluation in autonomous driving with (M)LLMs. The platform is open-source and is expected to provide conditions for empowering future research on autonomous driving. The paper also presents a baseline (M)LLM-driven closed-loop framework with a memory mechanism that has been illustrated in experiments of various scenarios including intersections, roundabouts, and ramps.

\vspace{12pt}
\balance
\bibliographystyle{IEEEtran}
\bibliography{ref}

\begin{thebibliography}{10}
\providecommand{\url}[1]{#1}
\csname url@samestyle\endcsname
\providecommand{\newblock}{\relax}
\providecommand{\bibinfo}[2]{#2}
\providecommand{\BIBentrySTDinterwordspacing}{\spaceskip=0pt\relax}
\providecommand{\BIBentryALTinterwordstretchfactor}{4}
\providecommand{\BIBentryALTinterwordspacing}{\spaceskip=\fontdimen2\font plus
\BIBentryALTinterwordstretchfactor\fontdimen3\font minus \fontdimen4\font\relax}
\providecommand{\BIBforeignlanguage}[2]{{%
\expandafter\ifx\csname l@#1\endcsname\relax
\typeout{** WARNING: IEEEtran.bst: No hyphenation pattern has been}%
\typeout{** loaded for the language `#1'. Using the pattern for}%
\typeout{** the default language instead.}%
\else
\language=\csname l@#1\endcsname
\fi
#2}}
\providecommand{\BIBdecl}{\relax}
\BIBdecl

\bibitem{openai2023gpt4}
OpenAI, ``{GPT-4} technical report,'' \emph{arXiv preprint arXiv:2303.08774}, 2023.

\bibitem{touvron2023llama}
H.~Touvron, T.~Lavril, G.~Izacard, X.~Martinet, M.-A. Lachaux, T.~Lacroix, B.~Rozi{\`e}re, N.~Goyal, E.~Hambro, F.~Azhar \emph{et~al.}, ``Llama: Open and efficient foundation language models,'' \emph{arXiv preprint arXiv:2302.13971}, 2023.

\bibitem{chowdhery2022palm}
A.~Chowdhery, S.~Narang, J.~Devlin, M.~Bosma, G.~Mishra, A.~Roberts, P.~Barham, H.~W. Chung, C.~Sutton, S.~Gehrmann \emph{et~al.}, ``{PaLM}: Scaling language modeling with pathways,'' \emph{Journal of Machine Learning Research}, vol.~24, no. 240, pp. 1--113, 2023.

\bibitem{gpt4v}
OpenAI, ``{GPT-4V(ision)} system card,'' {https://openai.com/research/gpt-4v-system-card}, 2023.

\bibitem{li2023knowledgedriven}
X.~Li, Y.~Bai, P.~Cai, L.~Wen, D.~Fu, B.~Zhang, X.~Yang, X.~Cai, T.~Ma, J.~Guo, X.~Gao, M.~Dou, B.~Shi, Y.~Liu, L.~He, and Y.~Qiao, ``Towards knowledge-driven autonomous driving,'' \emph{arXiv preprint arXiv:2312.04316}, 2023.

\bibitem{cui2024survey}
C.~Cui, Y.~Ma, X.~Cao, W.~Ye, Y.~Zhou, K.~Liang, J.~Chen, J.~Lu, Z.~Yang, K.-D. Liao \emph{et~al.}, ``A survey on multimodal large language models for autonomous driving,'' in \emph{Proceedings of the IEEE/CVF Winter Conference on Applications of Computer Vision}, 2024, pp. 958--979.

\bibitem{huang2023applications}
Y.~Huang, Y.~Chen, and Z.~Li, ``Applications of large scale foundation models for autonomous driving,'' \emph{arXiv preprint arXiv:2311.12144}, 2023.

\bibitem{wen2023dilu}
L.~Wen, D.~Fu, X.~Li, X.~Cai, T.~Ma, P.~Cai, M.~Dou, B.~Shi, L.~He, and Y.~Qiao, ``Dilu: A knowledge-driven approach to autonomous driving with large language models,'' \emph{arXiv preprint arXiv:2309.16292}, 2023.

\bibitem{sharan2023llm}
S.~Sharan, F.~Pittaluga, M.~Chandraker \emph{et~al.}, ``Llm-assist: Enhancing closed-loop planning with language-based reasoning,'' \emph{arXiv preprint arXiv:2401.00125}, 2023.

\bibitem{wang2023drivemlm}
W.~Wang, J.~Xie, C.~Hu, H.~Zou, J.~Fan, W.~Tong, Y.~Wen, S.~Wu, H.~Deng, Z.~Li \emph{et~al.}, ``{DriveMLM}: Aligning multi-modal large language models with behavioral planning states for autonomous driving,'' \emph{arXiv preprint arXiv:2312.09245}, 2023.

\bibitem{jin2023surrealdriver}
Y.~Jin, X.~Shen, H.~Peng, X.~Liu, J.~Qin, J.~Li, J.~Xie, P.~Gao, G.~Zhou, and J.~Gong, ``{SurrealDriver}: Designing generative driver agent simulation framework in urban contexts based on large language model,'' \emph{arXiv preprint arXiv:2309.13193}, 2023.

\bibitem{ma2023lampilot}
Y.~Ma, C.~Cui, X.~Cao, W.~Ye, P.~Liu, J.~Lu, A.~Abdelraouf, R.~Gupta, K.~Han, A.~Bera \emph{et~al.}, ``{LaMPilot}: An open benchmark dataset for autonomous driving with language model programs,'' \emph{arXiv preprint arXiv:2312.04372}, 2023.

\bibitem{highway-env}
E.~Leurent, ``An environment for autonomous driving decision-making,'' {https://github.com/eleurent/highway-env}, 2018.

\bibitem{dosovitskiy2017carla}
A.~Dosovitskiy, G.~Ros, F.~Codevilla, A.~Lopez, and V.~Koltun, ``{CARLA}: An open urban driving simulator,'' in \emph{Conference on Robot Learning}.\hskip 1em plus 0.5em minus 0.4em\relax PMLR, 2017, pp. 1--16.

\bibitem{caesar2021nuplan}
H.~Caesar, J.~Kabzan, K.~S. Tan, W.~K. Fong, E.~Wolff, A.~Lang, L.~Fletcher, O.~Beijbom, and S.~Omari, ``{NuPlan}: A closed-loop {ML}-based planning benchmark for autonomous vehicles,'' \emph{arXiv preprint arXiv:2106.11810}, 2021.

\bibitem{fu2024drive}
D.~Fu, X.~Li, L.~Wen, M.~Dou, P.~Cai, B.~Shi, and Y.~Qiao, ``Drive like a human: Rethinking autonomous driving with large language models,'' in \emph{Proceedings of the IEEE/CVF Winter Conference on Applications of Computer Vision}, 2024, pp. 910--919.

\bibitem{shao2023lmdrive}
H.~Shao, Y.~Hu, L.~Wang, S.~L. Waslander, Y.~Liu, and H.~Li, ``{LMDrive}: Closed-loop end-to-end driving with large language models,'' \emph{arXiv preprint arXiv:2312.07488}, 2023.

\bibitem{wen2023limsim}
L.~Wen, D.~Fu, S.~Mao, P.~Cai, M.~Dou, and Y.~Li, ``{LimSim}: A long-term interactive multi-scenario traffic simulator,'' \emph{arXiv preprint arXiv:2307.06648}, 2023.

\bibitem{cui2024drive}
C.~Cui, Y.~Ma, X.~Cao, W.~Ye, and Z.~Wang, ``Drive as you speak: Enabling human-like interaction with large language models in autonomous vehicles,'' in \emph{Proceedings of the IEEE/CVF Winter Conference on Applications of Computer Vision}, 2024, pp. 902--909.

\bibitem{cui2023large}
C.~Cui, Z.~Yang, Y.~Zhou, Y.~Ma, J.~Lu, and Z.~Wang, ``Large language models for autonomous driving: Real-world experiments,'' \emph{arXiv preprint arXiv:2312.09397}, 2023.

\bibitem{sha2023languagempc}
H.~Sha, Y.~Mu, Y.~Jiang, L.~Chen, C.~Xu, P.~Luo, S.~E. Li, M.~Tomizuka, W.~Zhan, and M.~Ding, ``{LanguageMPC}: Large language models as decision makers for autonomous driving,'' \emph{arXiv preprint arXiv:2310.03026}, 2023.

\bibitem{mao2023gpt}
J.~Mao, Y.~Qian, H.~Zhao, and Y.~Wang, ``{GPT-Driver}: Learning to drive with {GPT},'' \emph{arXiv preprint arXiv:2310.01415}, 2023.

\bibitem{wen2023road}
L.~Wen, X.~Yang, D.~Fu, X.~Wang, P.~Cai, X.~Li, T.~Ma, Y.~Li, L.~Xu, D.~Shang \emph{et~al.}, ``On the road with {GPT-4V (ision)}: Early explorations of visual-language model on autonomous driving,'' \emph{arXiv preprint arXiv:2311.05332}, 2023.

\bibitem{han2024dme}
W.~Han, D.~Guo, C.-Z. Xu, and J.~Shen, ``{DME-Driver}: Integrating human decision logic and {3D} scene perception in autonomous driving,'' \emph{arXiv preprint arXiv:2401.03641}, 2024.

\bibitem{liu2023visual}
H.~Liu, C.~Li, Q.~Wu, and Y.~J. Lee, ``Visual instruction tuning,'' \emph{arXiv preprint arXiv:2304.08485}, 2023.

\bibitem{ma2023dolphins}
Y.~Ma, Y.~Cao, J.~Sun, M.~Pavone, and C.~Xiao, ``Dolphins: Multimodal language model for driving,'' \emph{arXiv preprint arXiv:2312.00438}, 2023.

\bibitem{hdd}
V.~Ramanishka, Y.-T. Chen, T.~Misu, and K.~Saenko, ``Toward driving scene understanding: A dataset for learning driver behavior and causal reasoning,'' in \emph{Proceedings of the IEEE Conference on Computer Vision and Pattern Recognition}, 2018, pp. 7699--7707.

\bibitem{nuscenes}
H.~Caesar, V.~Bankiti, A.~H. Lang, S.~Vora, V.~E. Liong, Q.~Xu, A.~Krishnan, Y.~Pan, G.~Baldan, and O.~Beijbom, ``{nuScenes}: A multimodal dataset for autonomous driving,'' in \emph{Proceedings of the IEEE/CVF Conference on Computer Vision and Pattern Recognition}, 2020, pp. 11\,621--11\,631.

\bibitem{had}
J.~Kim, T.~Misu, Y.-T. Chen, A.~Tawari, and J.~Canny, ``Grounding human-to-vehicle advice for self-driving vehicles,'' in \emph{Proceedings of the IEEE/CVF Conference on Computer Vision and Pattern Recognition}, 2019, pp. 10\,591--10\,599.

\bibitem{wu2023language}
D.~Wu, W.~Han, T.~Wang, Y.~Liu, X.~Zhang, and J.~Shen, ``Language prompt for autonomous driving,'' \emph{arXiv preprint arXiv:2309.04379}, 2023.

\bibitem{chen2023driving}
L.~Chen, O.~Sinavski, J.~H{\"u}nermann, A.~Karnsund, A.~J. Willmott, D.~Birch, D.~Maund, and J.~Shotton, ``Driving with {LLMs}: Fusing object-level vector modality for explainable autonomous driving,'' \emph{arXiv preprint arXiv:2310.01957}, 2023.

\bibitem{qian2023nuscenes}
T.~Qian, J.~Chen, L.~Zhuo, Y.~Jiao, and Y.-G. Jiang, ``Nuscenes-qa: A multi-modal visual question answering benchmark for autonomous driving scenario,'' \emph{arXiv preprint arXiv:2305.14836}, 2023.

\bibitem{sima2023drivelm}
C.~Sima, K.~Renz, K.~Chitta, L.~Chen, H.~Zhang, C.~Xie, P.~Luo, A.~Geiger, and H.~Li, ``{DriveLM}: Driving with graph visual question answering,'' \emph{arXiv preprint arXiv:2312.14150}, 2023.

\bibitem{yu2022autonomous}
B.~Yu, C.~Chen, J.~Tang, S.~Liu, and J.-L. Gaudiot, ``Autonomous vehicles digital twin: A practical paradigm for autonomous driving system development,'' \emph{Computer}, vol.~55, no.~9, pp. 26--34, 2022.

\bibitem{chen2023end}
L.~Chen, P.~Wu, K.~Chitta, B.~Jaeger, A.~Geiger, and H.~Li, ``End-to-end autonomous driving: Challenges and frontiers,'' \emph{arXiv preprint arXiv:2306.16927}, 2023.

\bibitem{yang2023unisim}
Z.~Yang, Y.~Chen, J.~Wang, S.~Manivasagam, W.-C. Ma, A.~J. Yang, and R.~Urtasun, ``{UniSim}: A neural closed-loop sensor simulator,'' in \emph{Proceedings of the IEEE/CVF Conference on Computer Vision and Pattern Recognition}, 2023, pp. 1389--1399.

\bibitem{bae2020self}
I.~Bae, J.~Moon, J.~Jhung, H.~Suk, T.~Kim, H.~Park, J.~Cha, J.~Kim, D.~Kim, and S.~Kim, ``Self-driving like a human driver instead of a robocar: Personalized comfortable driving experience for autonomous vehicles,'' \emph{arXiv preprint arXiv:2001.03908}, 2020.

\bibitem{wang2023driving}
Y.~Wang, J.~He, L.~Fan, H.~Li, Y.~Chen, and Z.~Zhang, ``Driving into the future: Multiview visual forecasting and planning with world model for autonomous driving,'' \emph{arXiv preprint arXiv:2311.17918}, 2023.

\end{thebibliography}

\balance
\end{document}